\documentclass{ws-ijns}
\usepackage{graphicx}
\usepackage{cite}
\usepackage{url}
\usepackage{xcolor}

\begin{document}

\catchline{0}{0}{2005}{}{}

\markboth{Mozos et al.}{Stress Detection using Wearable and Sociometric Sensors}

\title{STRESS DETECTION USING WEARABLE PHYSIOLOGICAL AND SOCIOMETRIC SENSORS
\footnote{ \bf{This is the accepted manuscript of the article published in International Journal of Neural Systems, 27, 2, 2017. The Version of Record is available at DOI: 10.1142/S0129065716500416}}
}

\author{Oscar Martinez Mozos\footnote{Corresponding author.}}
\address{Dept. of Electronics, Computer Technology and Projects, Polytechnic University of Cartagena, \\
Plaza del Hospital, n1, 30202, Cartagena, Spain\\
E-mail: omozos@gmail.com\\
www.upct.es}

\author{Virginia Sandulescu}
\address{Faculty of Automatic Control and Computer Science, Politehnica University of Bucharest, \\
313 Splaiul Independentei, Bucharest, 060042, Romania \\
E-mail: s\_virg@yahoo.com \\
www.upb.ro}

\author{Sally Andrews}
\address{Division of Psychology, Nottingham Trent University, \\
Burton Street, Nottingham, NG1 4BU, UK \\
E-mail: s.andrews@lincoln.ac.uk\\
www.ntu.ac.uk}

\author{David Ellis}
\address{Department of Psychology, Lancaster University, \\
Bailrigg, Lancaster, LA1 4YW, UK \\
E-mail: d.a.ellis@lancaster.ac.uk\\
www.lancaster.ac.uk}

\author{Nicola Bellotto}
\address{School of Computer Science, University of Lincoln, \\ 
Brayford Pool, Lincoln, LN67TS, UK\\
E-mail: nbellotto@lincoln.ac.uk\\
www.lincoln.ac.uk}

\author{Radu Dobrescu}
\address{Faculty of Automatic Control and Computer Science, Politehnica University of Bucharest, \\
313 Splaiul Independentei, Bucharest, 060042, Romania \\
E-mail: rd\_dobrescu@yahoo.com \\
www.upb.ro}

\author{Jose Manuel Ferrandez}
\address{Dept. of Electronics, Computer Technology and Projects, Polytechnic University of Cartagena, \\
Plaza del Hospital, n1, 30202, Cartagena, Spain\\
E-mail: jm.ferrandez@upct.es\\
www.upct.es}

\maketitle

\begin{abstract}
Stress remains a significant social problem for individuals in modern societies. This paper presents a machine learning approach for the automatic detection of stress of people in a social situation by combining two sensor systems that capture physiological and social responses of people. We compared the performance of different Support Vector Machines and Boosting classifiers for this task. Our experimental results show that by combining the measurements from both sensor systems, we could accurately discriminate between stressful and neutral situations during a controlled Trier social stress test (TSST). Moreover, this paper assesses the discriminative ability of each sensor modality individually and considers their suitability for real time stress detection. Finally, we present an study of the most discriminative features for stress detection.
\end{abstract}


\begin{multicols}{2}

\section{Introduction}

While the world’s population continues to rise, the ratio of caregivers to those who require care is rapidly decreasing. New technologies, however, can offset this concern by automatically monitoring the physical and mental health of individuals across a variety of contexts. Specifically, stress remains a significant social problem for individuals and societies. Moreover, stressful situations encountered in everyday life can lead to additional negative mental states including depression and anxiety. While early detection could improve an individual's quality of life, the automatic detection of stress may also reveal key mechanisms and trigger points that underlie related negative behaviours and cognitions.

Stress is a natural reaction of the human body to an outside perturbing factor. Typical physiological responses include variations in heart rate, pulse, skin temperature, pupil dilation, and electro-dermal activity amongst others~\cite{Sloten2008,Renaud1997,Wikgren2012}. While small levels of stress may have beneficial effects on the body, stress often negatively impacts attention, memory, and decision-making~\cite{sandi2013,Strawski2006}. High levels of stress in the long-term also correlate with a variety of negative health outcomes including anxiety, depression and premature ageing~\cite{rai2011,Dickerson2004}. 

Psychosocial stress remains a large problem for society as a whole and has a detrimental impact on health care systems and the economies. According to the Mental Health Foundation in the UK~\cite{mhfuk}, around 12 million adults in the UK visit their general practitioner (GP) each year with mental health problems, many of which are related to or brought on by stress. As a consequence, 13.3 million working days are lost per year due to stress related illnesses. Moreover, the World Health Organization~\cite{who} recently calculated that stress costs around 8.4 million pound sterlings to UK enterprises. Therefore, treating negative mental states has become a priority in our societies, and numerous international organizations also consider stress to be a significant global problem, particularly within the work place~\cite{eu-osha,niosh,who-stress}. Unfortunately, current treatments offered by the National Health Service (NHS) in the UK (e.g. Cognitive Behavioural Therapy (CBT)~\cite{beck1975cbt}) have an average patient waiting time of 3-6 months~\cite{Perkins1994}. 

A related issue concerns future shortages of available health workers and doctors as reported by the Organisation for Economic Co-operation and Development (OECD)~\cite{oecd}. As a consequence of this shortage the ratio of caregivers to those who require advanced levels of care is rapidly decreasing. Therefore, there is a need to create new technologies to automatically monitor the physical and mental health of people in different situations, including activities of daily living and social interaction. These technologies, also known as Quality of Life Technologies (QoLTs), have emerged that aim to applying findings from different research areas that can together assist people during everyday activities, while simultaneously supporting health care providers. An emerging research topic inside QoLTs includes their potential application to stress and stress-realted illnesses~\cite{setz2010,Salahuddin2007embc,zhai2007embc,sun2012mobicase,Lu2012}. Thus, this paper was partly inspired by the need to create new wearable technologies that can monitor stress levels in people during their daily and social activities. 

In this paper, we aim to detect stressful behaviours by analysing measurements provided by several, non-invasive wearable sensors. In particular, we use a wearable sensor that provides real time physiological responses such as electrodermal activity and photoplethysmogram. In addition, we use a sociometric badge~\footnote{http://www.sociometricsolutions.com} to measure the social activity including body movement and voice. The measurements obtained from both sensors are processed and used as input to different classifiers that were trained to discriminate between stressful and neutral situations during a Trier Social Stress Test (TSST)~\cite{Kirschbaum1993}. In this work we use a binary classification and do not differentiate between different levels of stress. Our results show that the combination of physiological and activity measurements is able to discriminate with a high level of confidence between stressful and neutral situations on people while engaging across several activities. The main novelty of our systems is the combination of wearable sensors that makes the system a realistic option for measuring stress in social situations. As far as we now this is the first time that these sensor modalities are combined to detect stress. Finally we present a study of the most discriminative features obtained from the different sensors.

In this work we train a personal classifier for each participant because when consulting different psychologists they agreed on the fact that the response to stress is often very between individuals. Therefore, we opted for personalised systems.

The remainder of this paper is organised as follows. After considering some related work in Sect.~\ref{sec:rel_work}, we introduce the physiological wearable sensor and the sociometric badge in Sect.~\ref{sec:sensors}. In Sect.~\ref{sec:TSST} we describe our experimental setup. The process for data collection is explained in Sect.~\ref{sec:data_collec}, and our classification method is introduced in Sect.~\ref{sec:class}.  Experimental results on stress detection are presented in Sect.~\ref{sec:exp}. Finally, we conclude in Sect.~\ref{sec:conclusion}.

\section{Related Work}
\label{sec:rel_work}

Research concerning the automatic monitoring of mental states has grown exponentially during the last decade, and the resulting technologies aim to help patients monitor their own conditions while also supporting carer-providers. Examples include the detection and monitoring of stress~\cite{setz2010,Salahuddin2007embc,zhai2007embc,sun2012mobicase,Lu2012}, classification of different emotional states~\cite{Katsis2006,kim2004,chanel2011,Busso2004}, depression monitoring~\cite{sturim2011interspeech,adeli2015,Joshi2013depression}, obsessive compulsive disorder~\cite{Aydin2015ijns}, behaviour classification~\cite{Patel2010ijns}, cardiac states~\cite{Sankari2011,Martis2014cbm,Martis2014bspc}.



Various systems have been proposed for stress monitoring and/or detection using different physiological sensors. For example, the work reported in~\cite{setz2010} uses a wearable sensor to record electrodermal activity (EDA). This work applies the Montreal Imaging Stress Task (MIST)~\cite{Dedovic2005jpn} to induce states of stress on participants, and the results in indicate that EDA measurements can be used to discriminate stress from cognitive loads with 82.8\% accuracy. Although this work uses cross-validation for reporting results but does not present results of individual classifiers for each participant. In contrast, we study the personalization of the detector for each participant.   

Similarly, a mobile elecrocardiogram (ECG) sensor was used in~\cite{Salahuddin2007embc} to monitor stress by analysing ultra short term heart rate variability features during a Stroop test~\cite{stroop}. Authors conclude that ultra short term features could be used for stress monitoring, but they do not apply any machine learning classification tool to support this claim. In comparison, we present results by applying classifiers to the collected data.   

In~\cite{lee2004embc}, the authors analyzed EDA measurements together with skin temperature to discriminate between stressed and unstressed situations. Authors report correct classification rates of 96.6\% in the test data. However, the paper does not detail the final applied classification procedure.  
   
Finally, the work in~\cite{zhai2007embc} combines EDA signals, blood volume pulse, pupil diameter, and skin temperature to detect stress in participants while they took part in a modified Stroop test. The results report classifications of 90.1\% accuracy. However, this system seems difficult to implement as a daily life wearable technology due to the measurement of pupil diameter. In contrast, our proposed system is easier to wear by a patient.

The above mentioned works use physiological measurements in isolation and do not include social information. In contrast, we present a multi-modal system for stress monitoring that includes a sociometric sensor, that record activity of people and thus increases final detection rates. In addition, we focus on personalized systems by analysing individual results for each participant. Moreover, our experimental setup is based on the TSST which, rather than generate a stressful situation using a computer-based task, provides neutral and stressful conditions in a social setting. 

Voice is  also used as indicator for stress. The {\it StressSense} system introduced in~\cite{Lu2012} is a voice based stress detection mobile app that uses microphones to record the voice of the participant and classifies the obtained features using Gaussian Mixture Models. This work reports classification accuracies of 81\% and 76\% for indoor and outdoor environments, respectively. In our research, we also use the voice as one of the modalities for stress detection, but we combine it with other physiological and activity measurements, thus increasing the individual classification results.


Multi-modal approaches for stress detection have also been developed where activity and voice signals are added to physiological measurements in order to improve stress detection and monitoring. 

The work in~\cite{sun2012mobicase} presents a activity-aware mental stress detection scheme that combines Electrocardiogram (ECG), galvanic skin response (GSR), and accelerometer measurements from participants across three activities, i.e. sitting, standing, and walking, while users were subjected to mental stressors. This work applies the Stroop test~\cite{stroop} to induce stress. In our work, however, we are interested in detecting stress in social situations, therefore we apply the TSST to our participants. In addition, we use the sociometric badge to increase the modalities for stress detection by including voice and body movement.

Authors in~\cite{liao2004cvpr} introduce a multi-modal approach in which a rich set of activity and physiological measurements are used to monitor stress. In addition, they use a combination of facial expressions, eye movements, head movements, heart rate, skin temperature, GSR, mouse finger pressure, behavioural data from user interaction activities with the computer, and performance measures. The obtained results show that the presented activity measurements strongly correlate with the task workload and ensure reliability for further classification approaches. However, the proposed system is thought to be implemented in a desktop-based workplace environment, which restricts the situations in which it can be tested. Our system, however, is based on wearable sensors that allow the participants to freely move during their activities.

In addition to stress, the work presented in~\cite{Sung2005phd} aimed to detect deception while playing a poker game by including measurements like  voice variation, skin conductance, and heart rate. The experimental results show that it is possible to develop simple linear models with high accuracy that can be used to identify stress and bluffing for real money no-limit Hold'em tournaments. However, this system has been tested only in poker games where spikes in stress levels are artificially high. In contrast, we are aimed to develop stress detectors in different social activities.

The {\it LiveNet} system presented in~\cite{livenet} is a complete wearable hardware and software system for long term monitoring which, according to the authors, has potential applications for monitoring soldiers and Parkinson's patients. It could also detetct epilepsy seizures. However, these applications remain under development. The LiveNet system can also include a sociometric badge similar to the one we use in our research, but no classification results are presented using that sensor.

Finally, in~\cite{sandulescu2015iwinac}, we applied the same wearable physiological sensor to detect stress in people. However, that work only reports on 5 participants. In contrast, in this work we combine the physiological signals with the activity signals from the sociometric sensor and thus we improve classification results. In addition we use 18 patients in our experiments thus providing better reliability in the results.  

One of the main contributions of our work is the use of the sociometric badge for stress detection. This badge provides a way of capturing unconscious {\it social signals}~\cite{pentland2088} that have recently provided researchers with a novel and indirect way of capturing an individual's thoughts and cognitive states. Methods for assessing social signals also offer new tools for measuring levels of stress, particularly where verbal or written reports concerning underlying cognitive states may be incomplete or inaccurate. Therefore, we think the sociometric badge can be a powerful tool to monitor stress.

\section{Sensor Modalities}
\label{sec:sensors}

In this section we describe the two sensor devices used during our experiments and the features extracted from them.

To measure the physiological signals we use a wireless sensor~\cite{biopac} that is worn as a wristband is worn as a wristband (Fig.~\ref{fig:sensors} left) on the non-dominant hand of a subject and it is equipped with a set of electrodes situated on the fingers. This sensor connects wireless to a computer through a communication station. This setup allows the participant wearing the sensor to move freely during the experiments while signals are sent wirelessly to a computer. 

\begin{figurehere}
\begin{center}
\includegraphics[height=3.5cm]{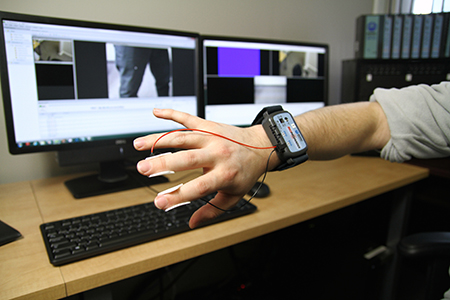} 
\includegraphics[height=3.5cm]{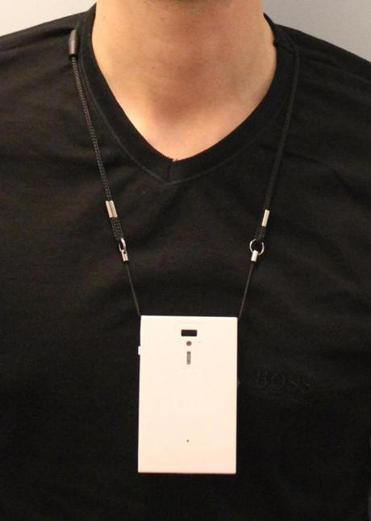}
\caption{The left picture shows the wireless sensor worn as a wristband. The right image depicts the sociometric sensor worn as a conference badge.}
\label{fig:sensors}
\end{center}
\end{figurehere}




From th electrodes situated on the fingers we obtain the three different measurements: the electro-dermal activity (EDA), the photoplethysmogram (PPG), and heart rate variability (HRV). The EDA signal, also called, skin conductance activity or galvanic skin response, is an indication of perspiration. A transient increase on the EDA signal is proportional to sweat secretion~\cite{darrow1964} and it is related to stress~\cite{Boucsein1992}. The PPG signal, also termed blood volume pulse, is obtained using a pulse oxiometer which illuminates the skin and measures the differences in light absorption. The amount of light that returns to the PPG sensor is proportional to the volume of blood in the tissue~\cite{peper2007}. Finally, the HRV signal represents the beat-to-beat variability over a given period of time and is computed by calculating the standard deviation of the average of normal-to-normal heartbeats~\cite{peper2007}. 

During our experiments, the EDA and PPG physiological signals were acquired at a 1000 Hz sampling frequency. Following  acquisition, the signals were down-sampled to 10 Hz. A filtering and artefact removal approach was also applied by using the routines included in the AcqKnowledge software \cite{biopac}. Specifically, the PPG signal was filtered using a band pass filter, with a low frequency of 0.5 Hz and a high frequency of 35 Hz. The EDA signal was filtered using a low pass filter with the cut-off frequency of 0.5 Hz. We obtain templates for the PPG signal by applying autocorrelation. The final list of recorded physiological features and their corresponding description are provided in Table~\ref{tab:features_phy}.

\begin{tablehere}
\tbl{List of physiological features from the Biopac wearable sensor.\label{tab:features_phy}}
{\begin{tabular}{@{}ll@{}}
\toprule
Feature & Description \\ \colrule
$eda$ & raw electro-dermal activity (EDA).\\
$eda_f$ & filtered EDA. \\
$ppg$ & raw pulse plethysmograph (PPG). \\
$ppg_t$ & PPG template using autocorrelation. \\
$hrv$ & heart rate variability (HRV). \\
\botrule
\end{tabular}}
\end{tablehere}


The second sensor we use in our system is the sociometric sensor~\cite{Olguin2008}, a small device that is worn around the neck like a conference badge (Fig.~\ref{fig:sensors} right). The sociometric badge is equipped with a microphone to record speech, an accelerometer to measure degree and direction of people's movement, a Bluetooth transmitter to measure the proximity of other sensors, and an infrared transmitter to measure when two sensor wearers are facing one another. In this work we only used the measurements provided by the microphone and the accelerometer. The data recorded from the sociometric badge can be translated into useful measures of social behaviour~\cite{pentland2088}.




In our experiments we recorded the signals from the sociometric badge at 10Hz. The samples were stored on the badge's internal memory and downloaded to the computer in an off-line process. The list of features that we used in our system based on the measurements provided by the sociometric badge is as follows: 

\begin{enumerate}

	\item Body movement ($bm$): the normalized acceleration magnitude over the 3 axis of movement.

	\item Body movement activity ($bm_{act}$): absolute value of the first derivative of the accelerometer's energy.

	\item Body movement rate ($bm_r$): second derivative of energy. It indicates the direction of change in someone's activity level.

	\item Posture activity ($pos_{act}$): absolute angular velocity.

	\item Posture rate ($pos_r$): angular acceleration. 

	\item Posture left right ($pos_{lr}$): orientation angle of the badge in left-right plane. 

	\item Posture front back ($pos_{fb}$): orientation angle of the badge in front-back plane.

	\item Speak ($voiced$): takes values 0 when the person is not speaking, and 1 when the person is speaking. 

	\item Silence ($unvoiced$): takes values 1 when the person is speaking, and 0 when the person is speaking. 

	\item Speech volume front ($vol_f$): average absolute value of amplitude of the front microphone.

	\item Speech volume back ($vol_b$): average absolute value of amplitude of the back microphone. 

	\item Volume consistency front ($volc_f$): measurement of change in speech volume.

	\item Volume consistency back ($volc_b$): measurement of change in speech volume. 

	\item Frequency and amplitude front
($hz0_f$, $amp0_f$), ($hz1_f$, $amp1_f$), ($hz2_f$, $amp2_f$), ($hz3_f$, $amp3_f$):
pairs of (frequency, amplitude) for the 1st  / 2nd / 3rd / 4th strongest peak in the frequency spectrum.

	\item Frequency and amplitude back
($hz0_b$, $amp0_b$), ($hz1_b$, $amp1_b$), ($hz2_b$, $amp2_b$), ($hz3_b$, $amp3_b$):
pairs of (frequency, amplitude) for the 1st  / 2nd / 3rd / 4th strongest peak in the frequency spectrum.	

	\item Front pitch ($pitch_f$): pitch of the voice from the front mic correlated with the fundamental frequency of the voice signal.
	
	\item Back pitch ($pitch_b$): pitch of the voice from the front mic correlated with the fundamental frequency of the voice signal:

\end{enumerate}

More detailed descriptions of the previous features can be found in~\cite{Olguin2008,Olguin2009,Olguin2010}.

\section{Experimental Protocol}
\label{sec:TSST}

To check the validity of our stress detector system we prepared an experimental setup where participants experienced different stressful situations. The work in~\cite{Dickerson2004} reviews more than 200 stress experiments, and concludes that the most effective tasks for inducing stress include public speaking and cognitive tasks. Therefore, our final design is based on the Trier Social Stress Test (TSST) \cite{Kirschbaum1993}, which includes both public speaking and cognitive tasks that place participants under high cognitive load.

The TSST is a very popular controlled experimental set-up and it has been used in more than 4000 sessions during the last few decades \cite{Dickerson2004}. This test consists of a neutral task followed by a public speaking task, a cognitive task, and a final neutral task. Each neutral task consists of 2 minutes of predefined standard questions including: ``How do you find the weather today?" and ``How did you get here?". The public speaking segment is a 5 minute interview for a desired job. After this, a cognitive task involves the participant counting back in steps of 13, starting from 1022.  All the previous tasks are performed in front of a live audience and video camera. The camera, however, is only used to increase stress levels~\cite{Dickerson2004} and recordings are not stored. The neutral tasks are considered as {\it non-stressful} situations, while the speaking and cognitive tasks are considered {\it stressful} conditions. The TSST experiment is used in this work as a valid proof of concept to show the capabilities our system to detect stress in social situations. 

In addition we ask the participants to fill in a State Trait Anxiety Inventory (STAI)~\cite{spielberger1983stai} to measure their levels of anxiety. The STAI questionnaire is divided into two sections. The state section contains 20 items that measure state anxiety, that is, how an individual feels right now; and the trait section contains a further 20 items that measure trait anxiety, that is, how an individual feels generally. Items are rated on a 4-point scale running from (1=‘Almost never’ to 4=‘Almost always’). The state section of the STAI was completed before and after the TSST session to ensure our protocol was having the desired effect on stress levels. 

Our protocol for the TSST was as follows. When a participant entered the room, she was given verbal and written information about the procedures involved in the experiment. The participant was then asked to fill in a consent form and to confirm that she did not suffer from any cardiovascular or anxiety disorder that might be affected by experiencing stress or that might affect the results. After being briefed, the participant was asked to fill in the STAI form. This provided an estimate current and general levels of stress. After fitted with both sensors, the participant was asked predefined neutral questions for 2 minutes in order to determine each individual's baseline measurements, which we used as a neutral state. Afterwards, the participant was asked to sit at a desk and prepare a presentation for a mock job interview for 3 minutes. They were provided with pen and paper. When 3 minutes of time expired, she was asked to hand back the sheet of paper, stand up in a predefined location inside the room, and begin her presentation. The participant was encouraged to speak continuously during 5 minutes. If the participant stopped during the presentation, at the first pause, she was told about the remaining time and asked to continue. At the next pause, she was asked a set of predefined typical interview questions including: ``What are your strengths/weaknesses?", ``Where do you see yourself in 5 years?" and so on. Following the presentation, the participant was asked to complete a cognitive task by counting backwards in steps of 13 from 1022 and a 5 minutes timer was started. If the participants made a mistake, she was asked to start the countdown again from the beginning (from 1022). At the end of this cognitive task, the participant was given a short time to relax while being debriefed. Another two minutes of neutral questions were then recorded. Finally, the participant was asked to re-complete the STAI questionnaire. A flowchart indicating the steps of our protocol is presented in Fig.~\ref{fig:tsst}. 

The times for each tasks during the TSST session are tentative and for most tasks, the duration of each differs by a few seconds between participants. 

\begin{figurehere}
	\begin{center}
	\includegraphics[width=0.9\columnwidth]{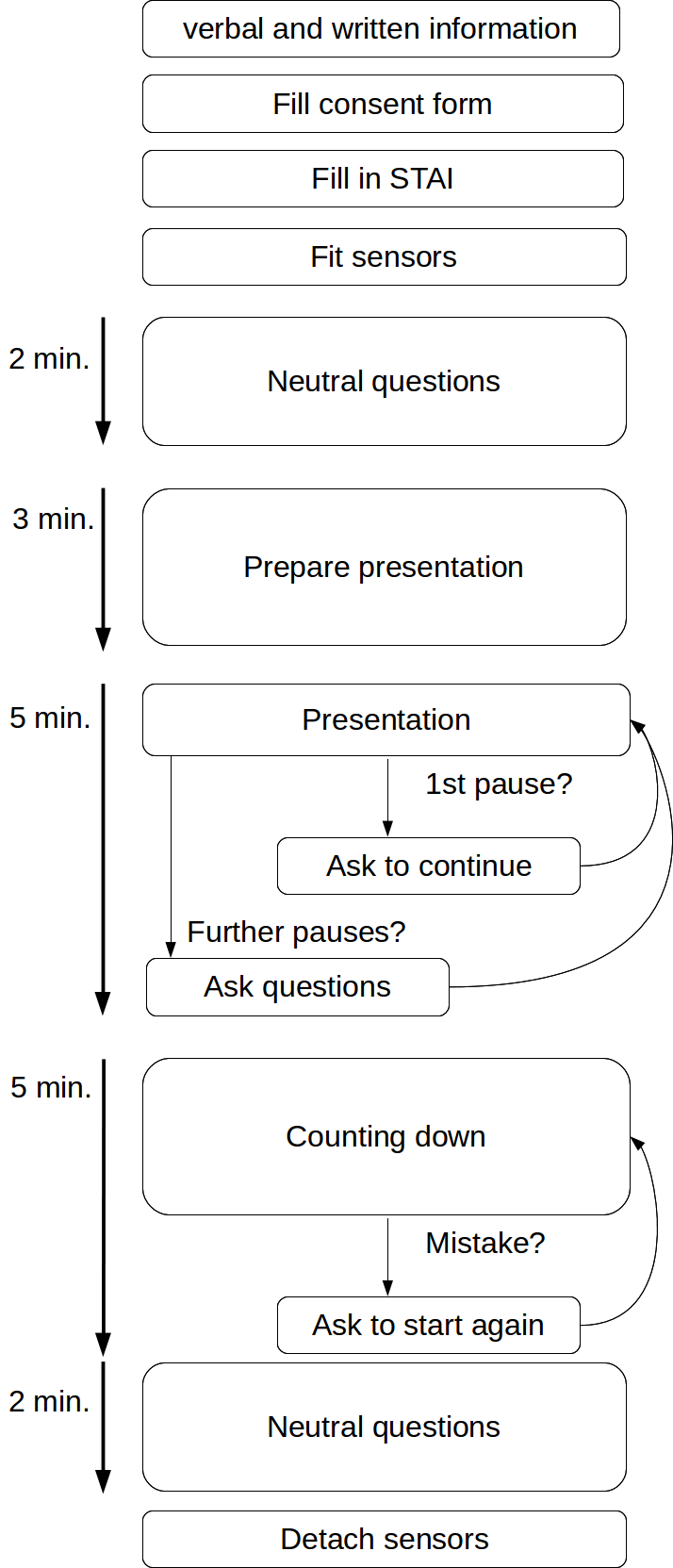}
	\caption{Flow chart of our TSST session.} 
	\label{fig:tsst}
\end{center}
\end{figurehere}

\section{Data Collection}
\label{sec:data_collec}

Eighteen participants $P_1-P_{18}$, who completed the TSST, were volunteering students from the School of Psychology at the University of Lincoln. The participants were aged 18 to 39, and included males and females. All participants signed a consent form before taking part. In addition, ethical approval for the experiment was obtained from the School of Psychology Research Ethics Committee at the University of Lincoln, following the British Psychological Society (BPS) ethical guidelines.

Each participant took part individually in one TSST session as introduced in Sec.~\ref{sec:TSST}. The participants wore both sensors at the same time during the TSST session, and the signals from both sensors were recorded and synchronised using the closest timestap. The final sample frequency for both sensor modalities was 10Hz. This value is below the maximum recommended for EDA measurements~\cite{Benedek2012}. The complete sets of signals for each sensor are listed in Sect.~\ref{sec:sensors}.  The final set of sampled signals for each time step t is thus defined as: 

\begin{equation}
	x^t = \{x_{phy}^t, x^t_{badge} \} ,
\end{equation}

\noindent where $x_{phy}^t$ is the vector containing the physiological signals as:

\begin{equation}
	x^t_{phy} = \{eda^t, eda^t_t, ppg^t, ppg^t_t, hrv^t \} ,
\end{equation}

\noindent and $x^t_{badge}$ is the vector containing the activity signals from the sociometric as:

\begin{align}
	x^t_{badge} = & \{ bm^t, bm^t_{act}, bm^t_r, pos^t{lr}, pos^t_{fb}, \nonumber \\
                & voiced^t, unvoiced^t, vol^t_f, vol^t_b, volc^t_f, volc^t_b, \nonumber \\
				& hz0^t_f, amp0^t_f, hz1^t_f, amp1^t_f, hz2^t_f, amp2^t_f, \nonumber \\
				& hz3^t_f, amp3^t_f, hz0^t_b, amp0^t_b, hz1^t_b, amp1^t_b, \nonumber \\ 
				& hz2^t_b, amp2^t_b, hz3^t_b, amp3^t_b, pitch_f, pitch_b \} .
\end{align}

The synchronised signals recorded from each participant $P_k$ were stored in the corresponding dataset $D_k$. The total number of synchronized samples for each participant $P_k$ is shown in Table~\ref{tab:data_size}. As explained in Sect.~\ref{sec:TSST}, our TSST sessions are composed of stressful and neutral scenarios. Therefore, each entry $x_t$ in the dataset $D_k$ is labeled as {\it stressed}, or {\it neutral} depending on the corresponding task (c.f. Sect.~\ref{sec:class}), i.e. we assume the participant is not stressed during the neutral activities as discussed in Sect.~\ref{sec:TSST}. Thus, the recorded dataset $D_k$ for each person was composed of the measurements obtained at each time interval $x_t$ together with their corresponding label as $D_k=\{(x_t,l_t)\}$, with $l_t \in L=\{stress, neutral\}$. The corresponding number of stressful and neutral samples in each dataset are also shown in Table~\ref{tab:data_size}. The average sample size for each patient is 13745 which is much higher than the dimension of the feature vector.

\begin{tablehere}
\tbl{Sample Size for Each Participant\label{tab:data_size}}
{\begin{tabular}{@{}lccc@{}}
\toprule
Participant & Total $|D_k|$ & Stress & Neutral \\ \colrule
$P_1$  & 12800 & 8200 & 4600\\
$P_2$  & 11620 & 8400 & 3220\\
$P_3$  & 13550 & 8350 & 5200\\
$P_4$  & 13450 & 8460 & 4990\\
$P_5$  & 13740 & 8760 & 4980\\
$P_6$  & 13000 & 8310 & 4690\\
$P_7$  & 15940 & 8600 & 7340\\
$P_8$  & 13610 & 7810 & 5800\\
$P_9$  & 12900 & 8420 & 4480\\
$P_{10}$  & 14200 & 8900 & 5300\\
$P_{11}$  & 15680 & 8660 & 7020\\
$P_{12}$  & 13530 & 8660 & 4870\\
$P_{13}$  & 14120 & 8560 & 5560 \\
$P_{14}$  & 13900 & 8810 & 5090\\
$P_{15}$  & 13350 & 8350 & 5000\\
$P_{16}$  & 14500 & 8800 & 5700\\
$P_{17}$  & 13480 & 8760 & 4720\\
$P_{18}$  & 14040 & 8820 & 5220\\
\colrule
Average $\pm$ std & 13745 $\pm$ 685  & 8535	$\pm$ 220 & 5210 $\pm$ 608 \\
\botrule
\end{tabular}}
\end{tablehere}

\begin{figure*}
	\begin{center}
	\includegraphics[width=\columnwidth]{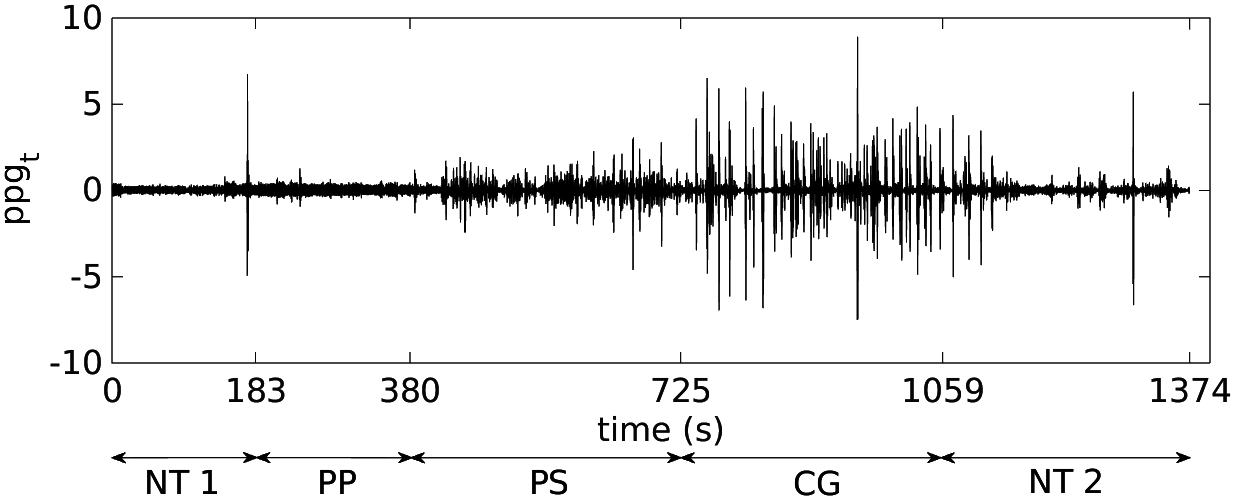} \hspace{2mm}
	\includegraphics[width=\columnwidth]{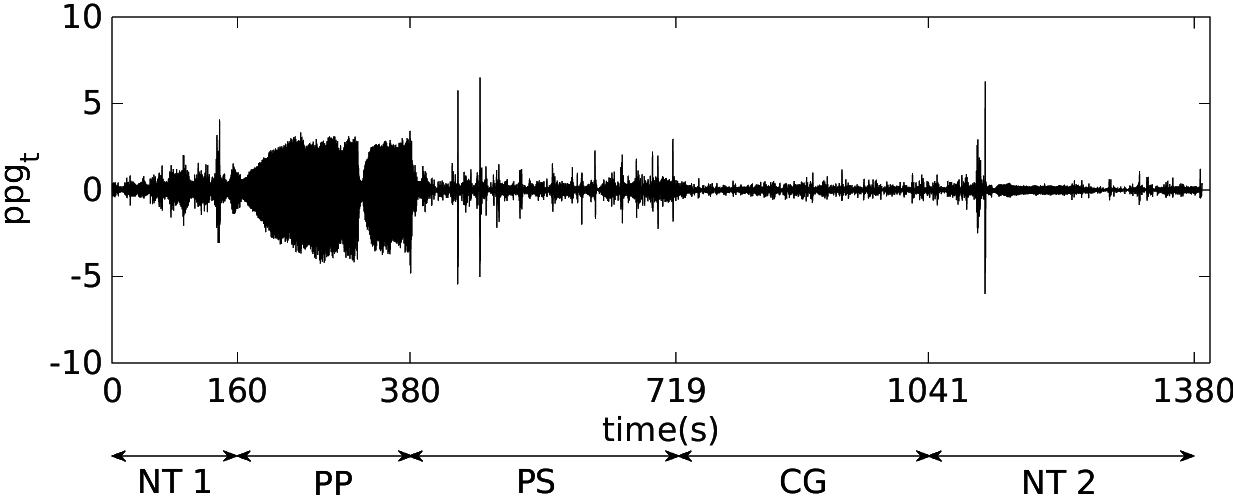} \\ \vspace{5mm}

	\includegraphics[width=\columnwidth]{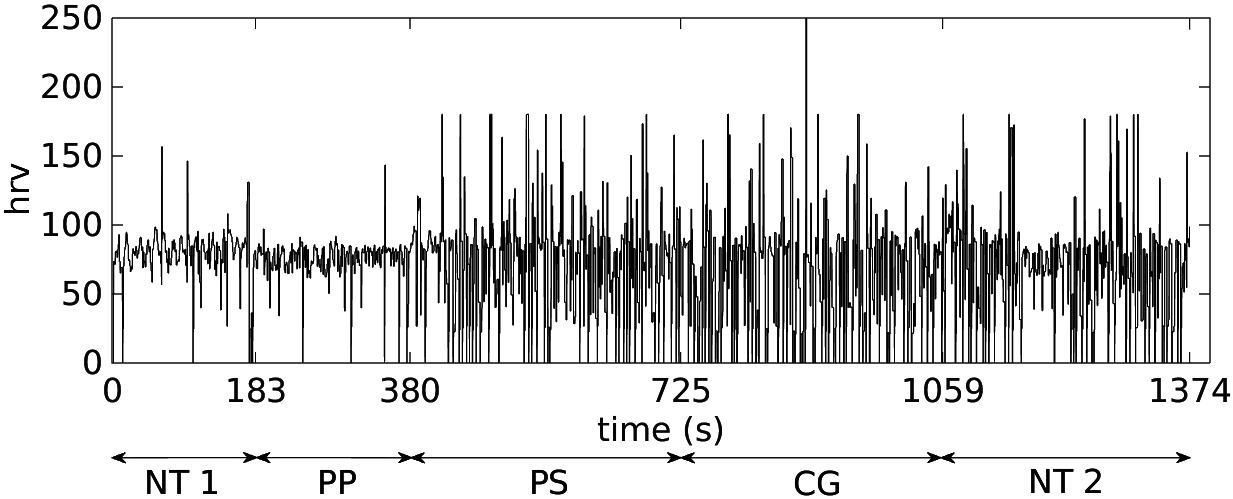} \hspace{2mm}
	\includegraphics[width=\columnwidth]{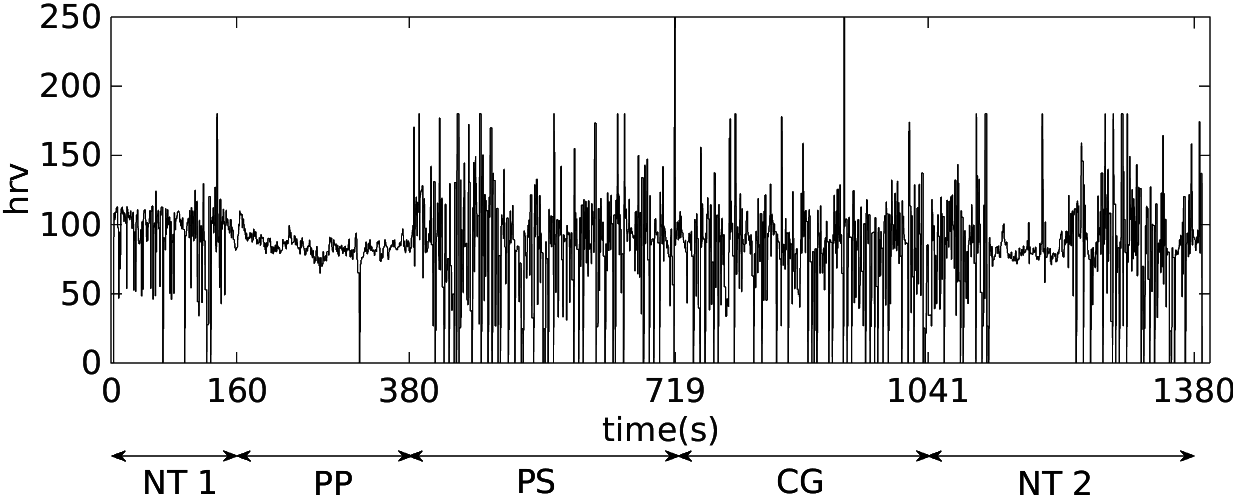} \\ \vspace{5mm}

	\includegraphics[width=\columnwidth]{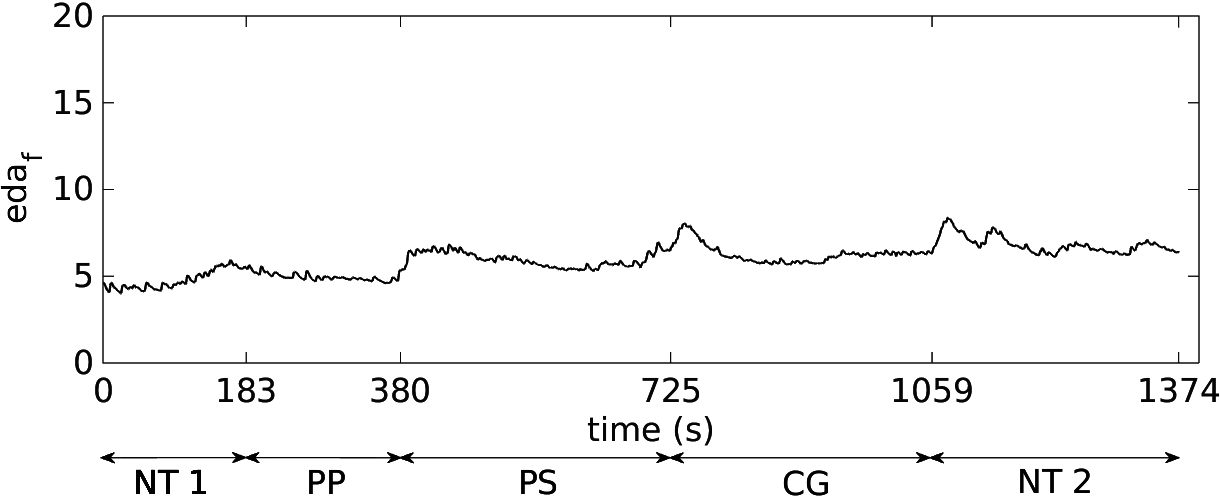} \hspace{2mm}
	\includegraphics[width=\columnwidth]{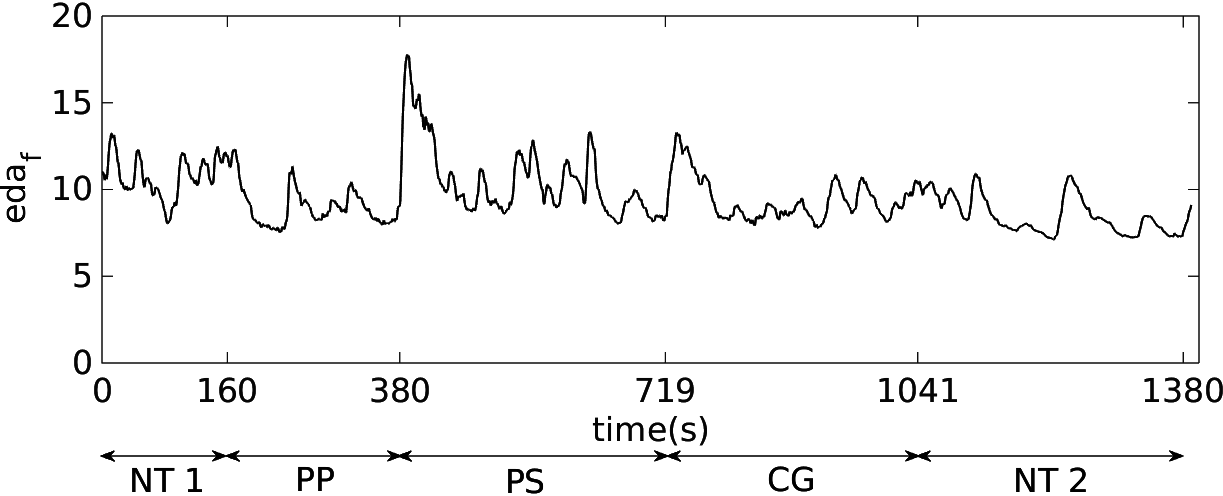} \\ \vspace{5mm}

	\includegraphics[width=\columnwidth]{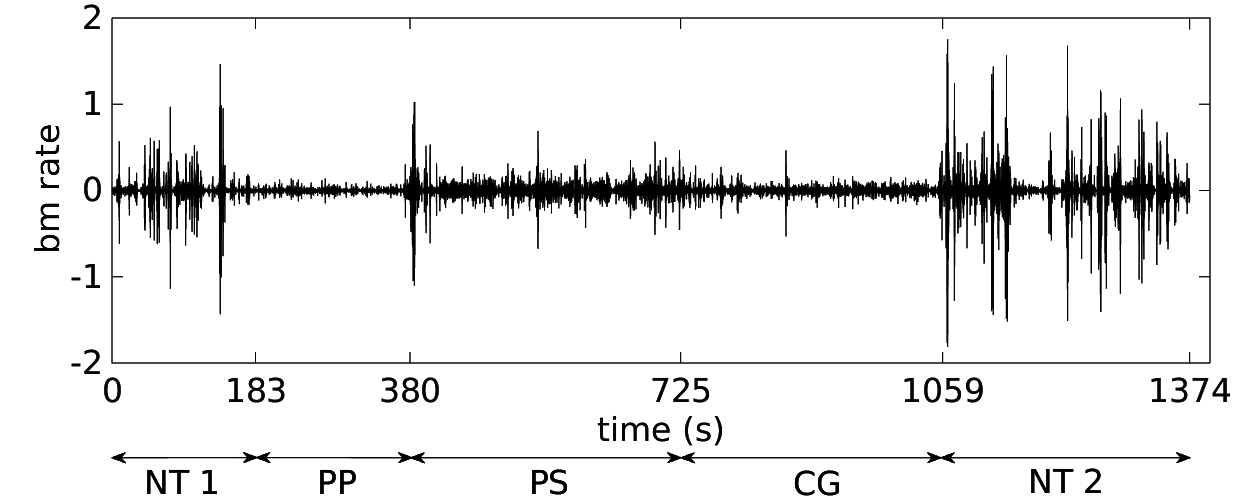} \hspace{2mm}
	\includegraphics[width=\columnwidth]{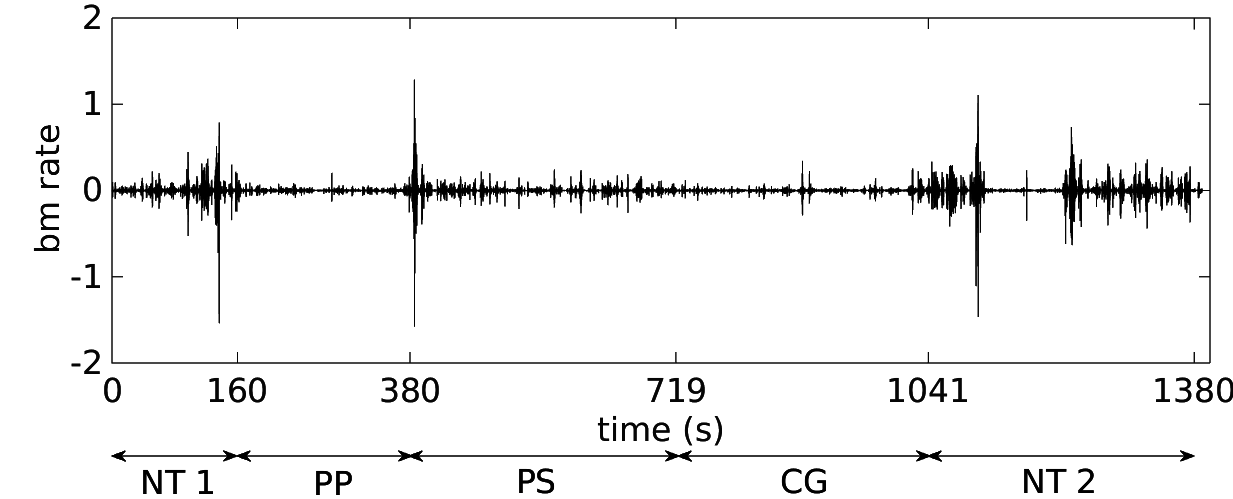}

	\caption{Example signals recorded during the TSST sessions for participant $P_{6}$ (left column), and participant $P_{16}$ (right column). The x-axis additionally presents the task the person was carrying out at each instant: NT 1 (First Neutral Task), PP (Presentation Preparation), PS (Public Speaking), Cognitive Task (CG), NT 2 (Second Neutral Task). }
	\label{fig:example_signals}
\end{center}
\end{figure*}

Some examples of the collected features during the TSST session are given in Figure~\ref{fig:example_signals}. In particular, we show the features PPG template ($ppg_t$), EDA ($eda_f$), HRV ($hrv$), and body movement rate ($bm_r$) for patients $P_{6}$ and $P_{16}$. The signals are shown with the corresponding task during the TSST session.

\section{Classification and Evaluation}
\label{sec:class}


In order to predict the state of each person at each point in time we compared different classification approaches. In particular we trained an SVM~\cite{Cortes1995ml,bishop2006book} and a classifier based on AdaBoost~\cite{Freund95Adaboost}. In this work we use a binary classification and do not differentiate between different levels of stress.

SVM-based predictors are popular in different areas~\cite{Chou2013cacie,zhang2015ijns,castillo2015ijns} including the classification mental states using physiological signals~\cite{sun2012mobicase,Joshi2013depression,zhai2007embc,Katsis2006,Aydin2015ijns,Patel2010ijns,Faust2015ijns,Rajendra2012ijns,Rajendra2013ijns}. SMVs were used in our experiments because the size of a trained SVM model is often much smaller than the volume of training data required in order to be successful. They can also be developed in real time.  In this work we compare two different kernel types. We first trained each SVM$_k$ using a radial basis function (RBF) kernel~\cite{adeli2000jte,karim2002jte,karim2003jte,karim2008jte} whose parameters $C$ and $\gamma$ were chosen using grid-search and cross-validation in the corresponding training data~\cite{Wei2010svm}. We also trained each SVM$_k$ using a linear kernel whose best parameter $C$ was selected empirically. For the training steps we used the software LIBSVM~\cite{CC01a} and LIBLINEAR~\cite{chang2008svmlinear}. 

AdaBoost is a popular boosting classifier applied in several areas~\cite{viola2001robust,martinez2005icra,arras2007icra} including biomedical, cognitive signals, and diagnosis~\cite{pineau2007,5309881,Cheng2011,1741-2552-1-4-004,GosztolyaBT13,Sabeti2009263,villar2015ijns}. Adaboost is a meta classifier that improves the classification capabilities of a weak classifier. In our case, we created a one-dimensional weak classifier from each feature as described in~\cite{martinez2005icra}. The only free parameter when training an AdaBoost classifier is the number $T$ of final weak classifiers that form the ensemble. This parameter is selected empirically in our experiments. When using single-feature weak classifiers, the output ensemble resulting from Adaboost can be used to analyse the features with best classification capabilities because the ensemble creates a ranking of weak classifiers according to their discrimination capabilities~\cite{viola2001robust,martinez2005icra,arras2007icra}. 




In this paper, we trained individual classifiers $C_k$ for each participant $P_k$ using the corresponding dataset $D_k$, and evaluated the classifier according to it. As introduced in the previous Sect.~\ref{sec:data_collec}, each dataset $D_k$ contains the synchronized physiological and activity signals for the corresponding participant $P_k$. For each individual $C_k$, we used $75\%$ of the corresponding dataset $D_k$ for training and the remaining $25\%$ for testing. To create these sets we used a random stratified selection to ensure the same class distribution in both the training and test sets. Afterwards, the attribute values of the training and test sets were scaled to the range $[-1 ,1]$. 

The classification results for each personalized classifier $C_k$ are evaluated using $accuracy$, $precision$, and $recall$, which are defined as:

\begin{equation}
	Accuracy=\frac{TP + TN}{TP + TN + FP + FN} 
\end{equation}

\begin{equation}
	Precision=\frac{TP}{TP + FP} 
\end{equation}

\noindent and finally $recall$ is given by:

\begin{equation}
	Recall=\frac{TP}{TP + FN} 
\end{equation}

\noindent where $TP$ indicates true positives, $TN$ true negatives, and  $FN$ false negatives. In our case we consider {\it positive} the examples labeled as $stress$. The values for $accuracy$, $precision$, and $recall$ lie in the range $[0,1]$ with values closer to 1 indicating better results.


\section{Experimental Results}
\label{sec:exp}

To ensure our approach for stress detection was valid, we ran a series of experiments that involved analysing the recorded physiological and activity signals of people under stress. We run one TSST for each participant, and each participant wore both sensors during the session.

We first present stress detection results when applying different classifiers to the synchronised physiological and activity signals. In addition, we analyse the capabilities of each individual sensor modality, and compare the classification results with those obtained in combination. Moreover, we analyse the discrimination capabilities of each feature. Finally, we discuss the results of the STAI questionnaires.


\subsection{Stress Detection using Physiological and Activity Sensors}
\label{subsec:combined}



In a first experiment we analysed the results of applying our proposed system using a combination of physiological and sociometric sensor. We trained an individual classifier $C_k$ for each participant as introduced in the previous Sect.~\ref{sec:class}.
We compare the classification results of three classifiers: SVM with a RBF kernel, SVM with linear kernel, and AdaBoost. For the RBF-based SVM we selected the best $C$ and $\gamma$  for each participant $P_k$ following the approach introduced in Sect.~\ref{sec:class}. For the linear SVM we found the value $C=10$ provided the best classification results in all participants. In the case of AdaBoost we empirically selected $T=300$.   

Table~\ref{tab:comp} shows the  classification results averaging over the 18 participants. As we can see from the table, in our experiments the AdaBoost classifier and the RBF-based SVM provide similar classification results, although AdaBoost provides slightly 
higher correct accuracy and precision rates. In both cases, the results are higher than in the linear SVM. We can conclude that both RBF kernel SVM and AdaBoost are suitable for the classification of stress, although AdaBoost presents a more straightforward training approach since no individual parameters need to be trained for each personalised classifier.  

\begin{tablehere}
\tbl{Comparison of RBF and linear kernels for the combined sensors.\label{tab:comp}}
{\begin{tabular}{@{}lccc@{}}
\toprule
Method & Accuracy & Precision & Recall\\ \colrule
AdaBoost (T=300) & 0.94$\pm$0.03 & 0.94$\pm$0.03 & 0.96$\pm$0.02 \\
RBF kernel SVM &  0.93$\pm$0.03 & 0.93$\pm$0.03 & 0.96$\pm$0.01 \\
Linear kernel SVM  &  0.85$\pm$0.04 & 0.84$\pm$0.04 & 0.94$\pm$0.02 \\
\botrule
\end{tabular}}
\end{tablehere}

In Table~\ref{tab:acc} we present in more detail the classification results of the AdaBoost classifier for each participant. Values for accuracy, precision and recall are similar for all the participant indicating that our personalised classification approach is suitable for stress detection with high confidence when applied to different individuals.

\begin{tablehere}
\tbl{Classification Results using Combined Sensors and AdaBoost\label{tab:acc}}
{\begin{tabular}{@{}cccc@{}}
\toprule
Participant & Accuracy & Precision & Recall\\ \colrule
$P_1$  & 0.95 & 0.95 &    0.97\\
$P_2$  &  0.91 &   0.92 &    0.96 \\
$P_3$  &  0.87 &   0.89 &    0.92\\
$P_4$  &  0.96 &  0.96  &  0.98\\
$P_5$  &  0.99 &   0.99&    0.99\\
$P_6$  &  0.89 &   0.90&    0.93\\
$P_7$  &  0.966 &   0.97 &    0.966\\
$P_8$  &  0.93 &   0.94 &    0.94\\
$P_9$  &  0.93 &   0.94 &    0.95\\
$P_{10}$  & 0.92 &  0.92 &    0.95\\
$P_{11}$  & 0.95 &   0.95 &    0.96\\
$P_{12}$ & 0.96 &   0.96 &    0.97 \\
$P_{13}$  & 0.91 &   0.92 &    0.94\\
$P_{14}$  & 0.97 &   0.96 &    0.98\\
$P_{15}$  & 0.91 &   0.90 &    0.95\\
$P_{16}$  & 0.95 &   0.96 &    0.96\\
$P_{17}$  & 0.98 &   0.98 &    0.99\\
$P_{18}$  & 0.96 &   0.96 &    0.97\\
\botrule
\end{tabular}}
\end{tablehere}


In addition, we include the average confusion matrix when using the Adaboost classifier in Table~\ref{tab:conf_ave}. The rows indicate the original label of the examples in the test set and the columns indicate the predicted label by the classifier. We observe high classification results in the main diagonal (true classifications) of the confusion matrix, which indicates that the classifier correctly labels the test examples with high accuracy.

\begin{tablehere}
\tbl{Average Confusion Matrix for Combined Sensors using AdaBoost \label{tab:conf_ave}}
{\begin{tabular}{@{}llrr@{}}
\toprule
  &   & Predicted Label \\	
  & \% & $Stress$ & $Neutral$\\ \colrule
  Original & $Stress$ & {\bf 96.05} & 3.95\\
  Label & $Neutral$ & 9.00 & {\bf 91.00} \\ 
\botrule
\end{tabular}}
\end{tablehere}
 
To exemplify the behaviour of our classifier we show its prediction for participants $P_6$ and $P_{16}$ during a complete TSST session in Fig.~\ref{fig:io_both}. The continuous line indicates the right label for each time instant (1 for stress and -1 for no-stress), while red diamonds indicate the predicted values by the classifier. The plots show that most of the time the classifier predicts the correct state of the person during the TSST session. We can see in the plot several false alarms. We think this can be due to the fact that the changes between relaxed and stressed situations need some time and this may be a reason for some false alarms. We plan to study in more detail these transitions in a future work.

\begin{figurehere}
	\begin{center}
	\includegraphics[width=\columnwidth]{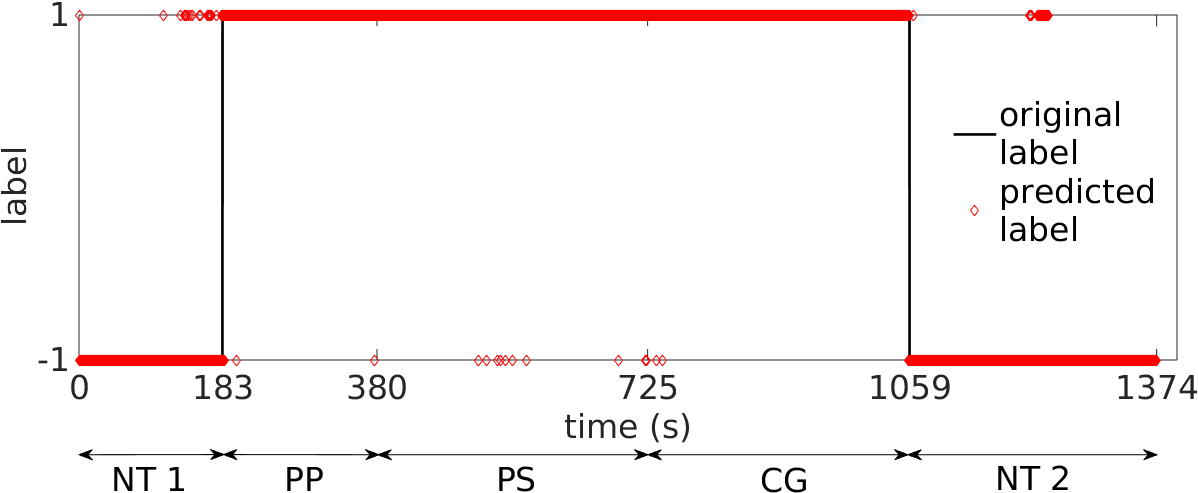} \\ \vspace{5mm}
	\includegraphics[width=\columnwidth]{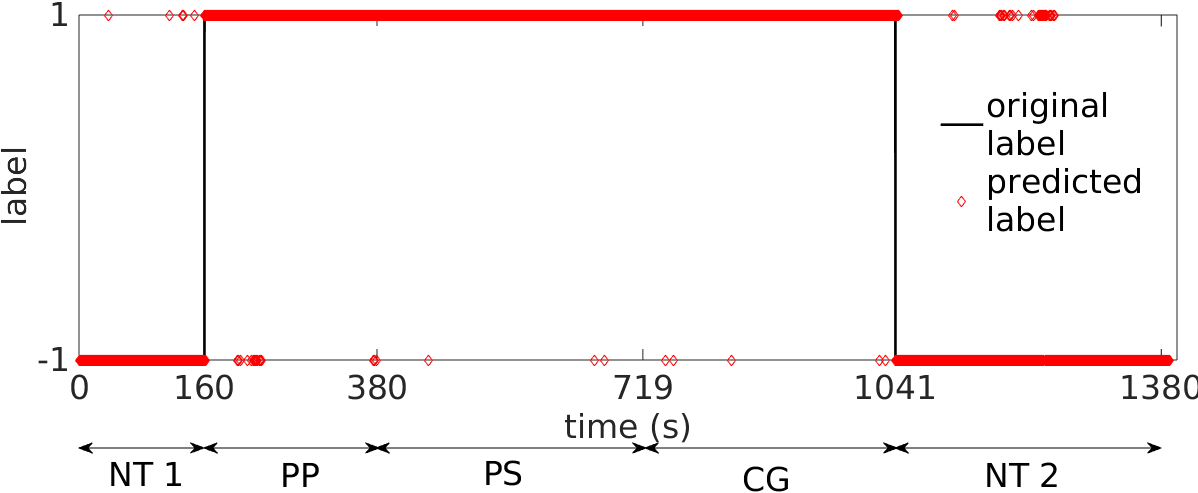}
	\caption{Classifier behaviour during a complete TSST session for participants $P_6$ (top) and $P_{16}$ (bottom) using both sensors. The y-axis represents the output of the classifier which is 1 for stressful tasks, and -1 for non-stressful tasks. The x-axis indicates each instant time in seconds during the TSST and the corresponding task: NT 1 (First Neutral Task), PP (Presentation Preparation), PS (Public Speaking), Cognitive Task (CG), NT 2 (Second Neutral Task). The continuous line indicates the right label for each time instant, and red diamonds indicate the predicted values from the classifier.}
	\label{fig:io_both}
\end{center}
\end{figurehere}

In addition we analyse the capability of each single sensor to detect stressful situations in the participants. In this way we can compare the detector capabilities of each sensor with respect the combination of both. We believe this information can be useful when deciding he sensor device to use in different situations. We include results using Adaboost since we showed in the previous section that this classifier provides the best classification results in our TSST sessions. For the physiological results we use as input for the classifier the feature vectors $x_{phy}$, whereas for the activity results we classify the vector $x_{badge}$ (see Sec.~\ref{sec:data_collec}). Table~\ref{tab:comp_mod} compares single sensor modalities and the combination of both. In all cases we have applied our AdaBoost classifier. As we can see in the table the combined modality provides better classification results. We also observe that the sociometric sensor provides better prediction results when used as single device. This may be due to the fact that the TSST session requires the participant to speak continuously and participants are also free to move during the experiments. A more detailed analysis of the features is provided in the following section.

\begin{tablehere}
\tbl{Comparison of single and combined modalities\label{tab:comp_mod}}
{\begin{tabular}{@{}lccc@{}}
\toprule
Method & Accuracy & Precision & Recall\\ 
\colrule
Physiological &  0.79$\pm$0.08  &  0.79$\pm$0.09  &  0.86$\pm$0.06 \\
Sociometric   &  0.89$\pm$0.03  &  0.90$\pm$0.03  &  0.92$\pm$0.03 \\
Physiological + Sociometric &  0.94$\pm$0.03  &  0.94$\pm$0.03  &  0.96$\pm$0.02 \\
\botrule
\end{tabular}}
\end{tablehere}

\subsection{Feature Analysis}
\label{sec:features}

As explained above, the AdaBoost algorithm creates a ranked ensemble of weak classifiers weighted by their discriminative capabilities. When each weak classifiers is constructed using a single feature, then the ranked ensemble is representative of the best features for the classification~\cite{viola2001robust,martinez2005icra,arras2007icra}.

In Table~\ref{tab:best} we indicate the first 5 most discriminative features that are selected by AdaBoost for each personalised classifier $C_k$ corresponding to patient $P_k$.
   
\begin{tablehere}
\tbl{Best 5 discriminative features for each $C_k$\label{tab:best}}
{\begin{tabular}{@{}lccccc@{}}
\toprule
Classifier & 1 & 2 & 3 & 4 & 5 \\ 
\colrule
$C_1$ & $hz3_f$ & $eda$ & $ppg_t$ & $amp0_f$ & $bm_{act}$ \\
$C_2$ & $hz3_f$ &	$pitch_f$	& $eda$	& $pos_{act}$ & $bm_{act}$ \\
$C_3$ & $pos_{act}$	& $pitch_b$	& $volc_b$	& $hz3_f$	& $volc_b$\\
$C_4$ & $hz3_f$	& $eda$	& $pitch_f$	& $pos_{act}$	& $pos_{act}$ \\
$C_5$ & $amp3_f$	& $eda$	& $pos_{lr}$	& $hz3_f$	& $eda$ \\
$C_6$ & $amp0_b$	&  $hz3_f$	&  $hz0_f$ & 	$bm_{act}$	& $eda$ \\
$C_7$ & $hz3_f$	& $eda$	&  $pitch_f$	&  $bm_{act}$	&  $pos_{act}$ \\
$C_8$ & $hz3_f$	&  $ppg$	& $amp3_f$	& $eda$	&  $volc_b$ \\
$C_9$ &  $amp3_f$	&  $bm_{act}$	&  $bm_{act}$	&  $amp1_b$	&  $ppg$ \\
$C_{10}$ & $bm_{act}$	&  $eda$	&  $pos_{act}$	&  $hz3_f$	&  $bm_{act}$ \\
$C_{11}$ & $hz3_f$	&  $bm_{act}$	&  $ppg$	&  $ppg_t$	&  $amp3_f$ \\
$C_{12}$ &  $amp3_f$	&  $pos_{act}$	&  $hz3_f$	&  $pitch_f$	&  $ppg$ \\
$C_{13}$ & $amp3_f$	&  $pos_{act}$	&  $eda$	&  $hz3_f$	&  $ppg$ \\
$C_{14}$ &  $amp3_f$	&  $hz3_b$	&  $bm_{act}$	&  $eda$	&  $pos_{act}$ \\
$C_{15}$ &  $vol_b$	&  $backp_v$	&  $eda$	&  $amp3_f$	&  $bm_{act}$ \\
$C_{16}$ &  $pos_{act}$	&  $eda$	&  $amp3_f$	&  $bm_{act}$	&  $pos_{act}$ \\
$C_{17}$ &  $pos_{act}$	&  $amp3_f$	 &  $eda$	&  $ppg_t$	&  $pos_{act}$ \\
$C_{18}$ &  $hz3_f$	&  $pitch_f$ &  $pos_{act}$	&  $amp3_f$	&  $amp3_f$ \\
\botrule
\end{tabular}}
\end{tablehere}

The most frequent features that appear in all the classifiers are $eda$ (16\% appearance), $pos_{act}$ (15\%), $hz3_f$ (14\%), $bm_{act}$ (13\%), and $amp3_f$ (13\%). The rest of features do not appear on the top five or they do with less than 10\%. To further check the discrimination power of the features from Table~\ref{tab:best} we carried out a classification experiment were we used our AdaBoost classifier with T=5, which selects only the first 5 most discriminative features (Table~\ref{tab:best}). The results of this experiment are shown in Table~\ref{tab:boost_best_5}.  We can see that using only the 5 most discriminative features we can obtain high classification results. This results indicate that most of the features obtained from the devices can be ignored in the classification and thus a more personalised system could be deployed for each patient.

\begin{tablehere}
\tbl{Classification using the 5 most discriminative features\label{tab:boost_best_5}}
{\begin{tabular}{@{}lccc@{}}
\toprule
Method & Accuracy & Precision & Recall\\ 
\colrule
AdaBoost (T=5)  &  0.83$\pm$0.04  &  0.84$\pm$0.07   &  0.88$\pm$0.04 \\
\botrule
\end{tabular}}
\end{tablehere}

\subsection{STAI Questionnaire Analysis}
\label{sec:stai}

We recorded participants’ self-reported anxiety levels using the STAI~\cite{spielberger1983stai} before and immediately after the TSST session (all Cronbach alphas $>0.9$~\cite{cronbach1951alpha}). A paired sample t-test demonstrated that participants felt more stressed following the TSST [t(34) = 3.54, p $>$ 0.01] suggesting that our procedure provided a stressful environment. While changes in state-levels of anxiety indicate that participants felt more stressed following a TSST session, it does not confirm how levels of stress varied across each part of the protocol independently.  During the TSST, we assume that participants became more relaxed during the neutral sections, but this is difficult to validate with 100\% accuracy because any additional measures of self-report would both disrupt the flow of the experiment and reduce the ecological validity of the procedure by reminding participants they were taking part in an experiment. Future research aims to explore these changes beyond a laboratory setting in conjunction with real-time self-report. 



\section{Conclusion}
\label{sec:conclusion}

In this paper we have presented a novel approach for stress detection using a combination of wearable physiological and sociometric sensors. The experiments were carried out under controlled conditions during different TSST sessions. Our wearable system allows the state of a participant to be determined at any instant by providing an accurate decision regarding his/her stress state at any time. Our classification results demonstrate that our method and analysis provides a useful tool for real-time stress detection. In the future this may allow other researchers to consider for example, the effect of real-time feedback and even reveal specific triggers that lead to high and unhealthy levels of stress. This will be essential when developing relevant early warning systems alongside future interventions. 

Although the TSST is a controlled setup and does not completely represent general everyday activities, it is used in our work because it is a well known method to create stress so we think the first step is to validate our system under a controlled situation so that the results are more reliable. Moreover, a personal interview is an example of a social activity that many people will have to face. However, future work will include longer experiments in order to use these technologies in daily life activities, although stressing people for longer periods of time may imply some ethical issues. 

The combined sensor solution presented in this paper may not be a perfect general solution for detecting stress on a day-to-day basis because the sensors can be uncomfortable in the long term, in particular the wireless physiological sensor, although the sociometric sensor can be worn as a simple badge during longer periods of time. However, one of the main contributions of this paper is to show the individual capabilities of each sensor modality to detect stress, and to present results that will aid in the future selection of appropriate sensors in different situations. The sociometric badge can be worn as part of an anyones daily activities (e.g. in an office environment) however, the physiological sensor may be more suitable within controlled environments (e.g. patients in hospital). In addition, new physiological and activity sensors continue to be developed whic are smaller and more ergonomic, and the results presented in this paper can easily transfer across. 

Classification results using only the sociometric badge indicate that easy to wear activity recording sensors are likely to be suitable for monitoring everyday stress. While the classification rates were lower when compared to a combination of both physiological and sociometric sensors, they still remained high. In addition, the sociometric badge allows for the recording of social interactions between participants. Thus, studying stress levels while interacting in a variety of social activities is another future area of research that we would like to explore in the future. Combined signals from social interactions could be combined with individual variation to increase the modalities when detecting stress in individuals and groups. 

Finally, similar sensors to those contained within the sociometric badge can be readily found in other devices including intelligent bracelets, smart phones or smart watches. Therefore, a further series of studies is likely to explore the usefulness of these devices alongside self-report when monitoring everyday levels of stress over longer periods of time that go beyond the psychological laboratory. 

\section{Acknowledgements}
This work was partially funded by a Sectoral Operational Programme Human Resources Development 2007-2013 from the Ministry of European Funds (POS- DRU/159/1.5/S/132397), and by a Research Investment Grant from the University of Lincoln (RIF2014-31).

\bibliographystyle{ws-ijns}
\bibliography{mozos2015ijns}

\end{multicols}
\end{document}